\documentclass[conference]{IEEEtran}
\IEEEoverridecommandlockouts
\usepackage{amsmath,amssymb,amsfonts}
\usepackage{algorithmic}
\usepackage{graphicx}
\usepackage{textcomp}
\usepackage{xcolor}
\usepackage{bbm}
\usepackage{caption}
\usepackage{lipsum}
\usepackage{hhline}
\usepackage{multirow}

\usepackage[sorting = none, backend = bibtex, style=numeric-comp]{biblatex}
\AtBeginBibliography{\normalsize}
\usepackage{setspace}
\DeclareRobustCommand*{\IEEEauthorrefmark}[1]{%
  \raisebox{0pt}[0pt][0pt]{\textsuperscript{\footnotesize #1}}%
}
\begin{document}

\title{SLPC: a VRNN-based approach for stochastic lidar prediction and completion in autonomous driving
}

\author{\IEEEauthorblockN{George Eskandar\IEEEauthorrefmark{1},
Alexander Braun\IEEEauthorrefmark{2},
Martin Meinke\IEEEauthorrefmark{2},
Karim Armanious\IEEEauthorrefmark{1} and
Bin Yang\IEEEauthorrefmark{1}}
\IEEEauthorblockA{\IEEEauthorrefmark{1}University of Stuttgart, Institute of Signal Processing and System Theory\\ Stuttgart, Germany\\}
\IEEEauthorblockA{\IEEEauthorrefmark{2}Robert Bosch GmbH, Lidar Perception\\ Stuttgart, Germany}}

\maketitle

\begin{abstract}
Predicting future 3D LiDAR pointclouds is a challenging task that is useful in many applications in autonomous driving such as trajectory prediction, pose forecasting and decision making. In this work, we propose a new LiDAR prediction framework that is based on generative models namely Variational Recurrent Neural Networks (VRNNs), titled Stochastic LiDAR Prediction and Completion (SLPC). Our algorithm is able to address the limitations of previous video prediction frameworks when dealing with sparse data by spatially inpainting the depth maps in the upcoming frames. Our contributions can thus be summarized as follows: we introduce the new task of predicting and completing depth maps from spatially sparse data, we present a sparse version of VRNNs and an effective self-supervised training method that does not require any labels. Experimental results illustrate the effectiveness of our framework in comparison to the state of the art methods in video prediction. 
\end{abstract}

\section{Introduction}

Autonomous cars continuously use depth information to capture the geometry of their environment. Many vision tasks such as object detection, scene flow, path planning and segmentation can be built on this input. \textbf{Li}ght \textbf{d}etection \textbf{a}nd \textbf{r}anging technology (LiDAR) is heavily integrated in the autonomous driving industry as a remote sensing method by virtue of its high accuracy and long range. However, in bad weather conditions such as heavy rain, snow, fog, low hanging clouds or high sun angles, LiDAR can suffer from reduced precision due to reflection and refraction of light beams, unlike radar sensors which are more robust in these situations \cite{rad3,rad4}. Additionally, many LiDAR sensors currently available on the market do have a limited number of horizontal scanning lines (e.g. 16, 64, 128) which provides sparse depth measurements, thus capturing the environmental geometry with a precision that can still be improved. As such, predicting LiDAR measurement of future frames is desirable not just in academic research but also in industrial applications as in autonomous driving. For instance, on a short timescale it can work as a fallback real-time system in case of critical weather conditions and/or sensor failure. 

Despite its potential, LiDAR pointcloud prediction poses several technical challenges. First, unlike images, LiDAR scans are sparse and do not present strong local features like edges and corners. Second, it's hard to simultaneously learn the depth information, the distribution of the points projected on a camera frame (or sparsity shape) and the distribution of possible outcomes. Thirdly, we want to predict LiDAR scans at a high resolution, which usually makes the prediction task more challenging because a larger pixel-space leads to more uncertainty in the prediction. Thus, our research lies at the intersection of video prediction and depth inpainting, also known as depth completion. 

The video prediction task is currently a hot topic in computer vision. Deep learning techniques have been recently used to predict future frames in videos. There have been two big approaches to video prediction: the first is to directly generate the pixels of the new frame like \cite{DBLP:journals/corr/LotterKC16, DBLP:journals/corr/KalchbrennerOSD16, DBLP:journals/corr/SrivastavaMS15, DBLP:journals/corr/OhGLLS15}. The second is to predict motion vectors and learn a transformation of the past pixels, like in \cite{DBLP:journals/corr/AgrawalCM15,DBLP:journals/corr/FinnGL16,DBLP:journals/corr/abs-1811-00684}. What these frameworks have in common is that they use a combination of convolutional (CNNs and 3D-CNNs) and recurrent neural networks (LSTMs, GRUs, convLSTMs) to predict subsequent measurements. More recent frameworks use generative approaches like VAEs and Generative Adversarial Networks (GANs) \cite{kingma2013auto,rezende2014stochastic,goodfellow2014generative,rad1,rad2} and learn a time-dependent multimodal probability distribution in pixel-space. In \cite{DBLP:journals/corr/ChungKDGCB15}, the Variational Recurrent Neural Network (VRNN) was introduced where a prior probability distribution is defined over latent random variables in the hidden state of an RNN. Sampling from the prior leads to different outcomes. State-of-the-art video prediction consists of variations of VRNNs like \cite{DBLP:journals/corr/abs-1802-07687,DBLP:journals/corr/abs-1804-01523,DBLP:journals/corr/abs-1904-12165}, as they model uncertainty in the prediction in contrast to the deterministic approaches. 

On the other hand, Depth Completion refers to the task of spatially inpainting sparse depth maps. Traditionally,  pointcloud is projected onto the camera frame resulting in an irregular depth map. Research in this field consists in developping network architectures which produce accurate dense depth maps \cite{DBLP:journals/corr/abs-1708-06500}. Recently, Patil et al \cite{Patil_2020} demonstrated that considering temporal correlation improves depth completion.
\begin{figure*}
    \centering
    \includegraphics[width=\textwidth]{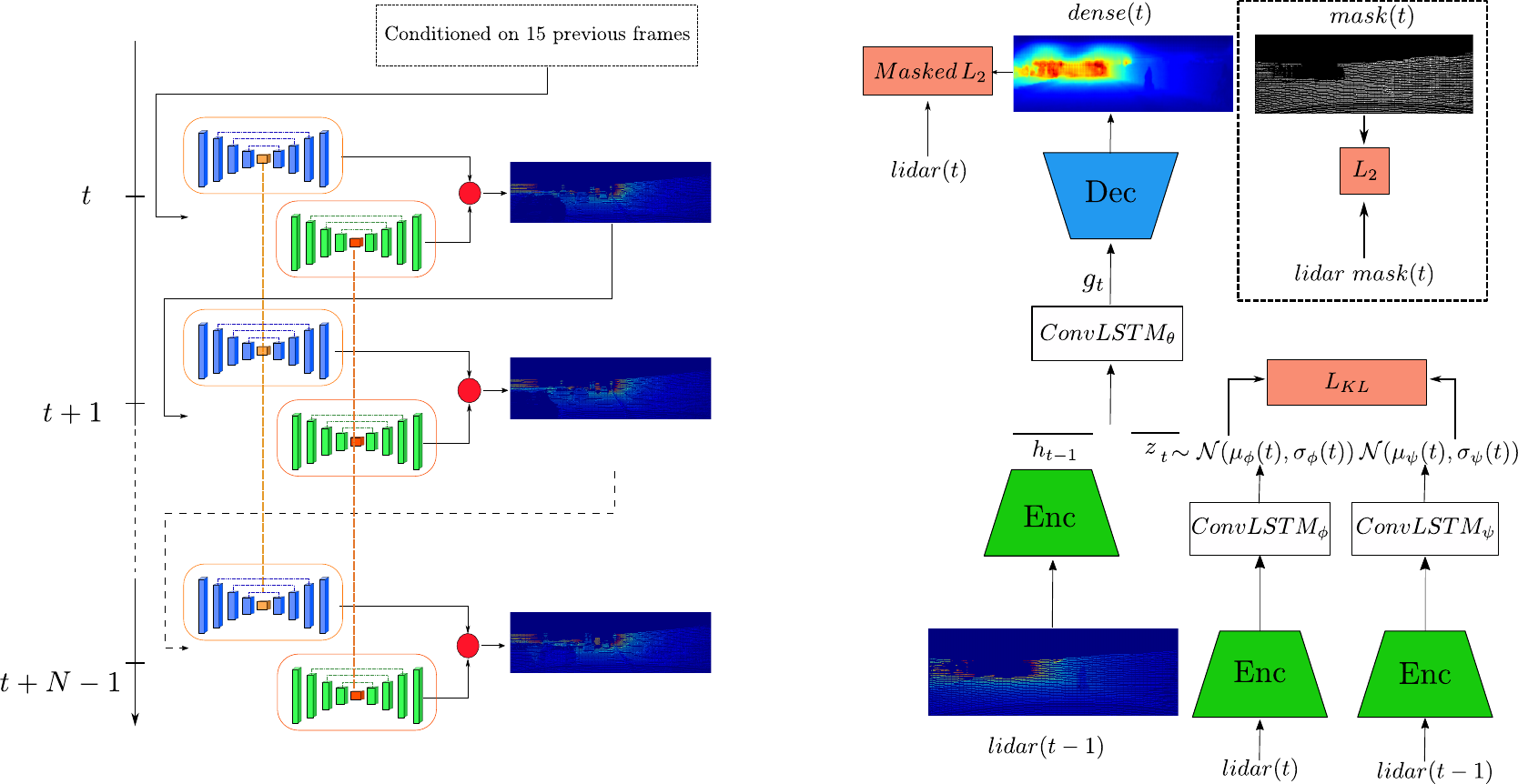}
    \caption{Left: In inference, the Depth Network predicts a more complete Depth map while the Mask network predicts a sparsity mask that downsamples the dense depth map. Right: Network architecture and losses. The two parallel networks, depth and mask, share the same architecture but use different loss functions as shown in the box with dashed boundaries. }
    \label{fig:twoNetworks}
\end{figure*}
In autonomous driving and robotics, the prediction of the future dynamic environment using deep learning has been investigated from two main aspects.  Some methods involve predicting a dynamic occupancy grid (a pixel is occupied if there is a rigid object or obstacle) like \cite{schreiber2019motion} and \cite{DBLP:journals/corr/abs-1904-12374}. On the other hand, some recent frameworks estimate absolute depth values like \cite{DBLP:journals/corr/abs-1902-06919} which predict subsequent LiDAR measurements for indoor robotic navigation. The most recent framework, Scene PointCloud Sequence Forecasting (SPCSF) \cite{weng2020unsupervised}, generates future LiDAR pointclouds in a determined way using the Chamfer Distance loss function \cite{DBLP:journals/corr/FanSG16}, by first predicting future range map frames and projecting them in 3D.

 In this work, we introduce the task of predicting dense depth maps from previous sparse LiDAR measurements. This would enable the generation of more reliable depth information in bad weather condition, by leveraging temporal correlation between past frames. We use the depth map representation, which projects the pointcloud in the front view of the car on the camera frame of reference, in contrast to the the range map representation in \cite{weng2020unsupervised}. This is better suited to autonomous driving pipelines which perform sensor fusion between RGB monocameras and LiDAR sensors. To tackle the challenge, our model extends the state-of-the-art of video prediction with a modified architecture, losses and a novel training setup, tailored to the sparse LiDAR maps. Not only do we predict future pointclouds, but we also inpaint them spatially to have dense depth maps that are richer in geometric information. A thorough quantitative and qualitative analysis is conducted between the proposed framework and the state-of-the-art baselines in both video and LiDAR prediction frameworks for single and multiple frame prediction.
\section{Methods}
\label{sec:methods}
In this section, we introduce the proposed SLPC framework for Stochastic LiDAR Prediction and Completion. This framework utilizes as baseline the Stochastic Video Prediction with Learned Prior SVG-LP model. However, we expand on the prior framework with a different architecture, loss functions and training methods that are tailored to the sparse nature of our data. This should enable the prediction of subsequent LiDAR frames in a more robust manner.
\vspace{-1mm}
\subsection{VRNN with learned prior}
\vspace{-1mm}
A VRNN is a variational RNN that generates future frames $Q_{t}$ based on the encoded previous frames $lidar_{1..t-1}$ and a latent variable $z_{t}$ sampled from a prior probability distribution $p(z)$. The latter encodes stochastic spatio-temporal information not captured by the deterministic encoder. In our baseline SVG-LP, this prior distribution, $q_{\psi}(z_{t}|x_{1..t-1})$, is chosen to be a conditional time-varying Gaussian distribution with fixed variance $\mathcal{N}(\mu_{\psi}(t), \sigma_{\psi}(t))$. During training, the prior is forced to be similar to the posterior distribution $q_{\phi}(z_{t}|x_{1..t}) = \mathcal{N}(\mu_{\phi}(t), \sigma_{\phi}(t))$ by minimizing the KL-divergence between them.
As  depicted in Fig. \ref{fig:twoNetworks}, the whole model thus consists of four parts:
\vspace{-0.5mm}
\begin{itemize}
    \item a frame encoder ($Enc$)  that encodes the previous frame $lidar_{t-1}$ into a variable in a latent space $h_{t-1}$.
    \item a prior/posterior model ($ConvLSTM_{\psi}$, $ConvLSTM_{\phi}$) maps the latent variable into a probability distribution $\mathcal{N}(\mu_{\psi}(t), \sigma_{\psi}(t))$ / $\mathcal{N}(\mu_{\phi}(t), \sigma_{\phi}(t))$.
    \item a frame predictor ($ConvLSTM_{\theta}$) encodes the combined $h_{t-1}$ and $z_{t} \sim \mathcal{N}(\mu(t), \sigma(t))$ into a state  $g_{t}$.
    \item a decoder ($Dec$) that generates $dense_{t}$ / $mask_{t}$ from $g_{t}$ and skip connections from encoder layers.
\end{itemize}

\subsection{The sparsity problem}
The sparsity problem of projected pointclouds is addressed in three ways: the architecture, losses and the training method.
Since our goal is to predict and complete a sequence of sparse LiDAR frames $\mathcal{\hat{Q}}(t)$, we formulate our problem as $\mathcal{\hat{Q}}(t) = dense(t) \odot mask(t)$, where $dense(t)$ is a dense depth map and $mask(t)$ is a mask that downsamples the dense map to obtain a sparse pointcloud. We use a divide-and-conquer approach and use  2 separate VRNNs to predict each of $dense(t+1)$ and $mask(t+1)$. The choice of this formulation allows for more accurate and complete depth maps all while having a sparse pointcloud that can then be fed back as input of the model to generate more frames in an autoregressive manner. \par   

The architecture and input of the $\mathbf{Depth}$ and $\mathbf{Mask}$ networks are the same. For more robustness against the sparse LiDAR, the encoder has three sparse convolution layers \cite{DBLP:journals/corr/abs-1708-06500}. The operation is a masked convolution that normalizes the output over the number of pixels having valid depth values, and it has been shown in \cite{DBLP:journals/corr/abs-1708-06500} to lead to less noisy results. The operation can be mathematically described in Eq. \eqref{sparseconv} as:
 \begin{equation}
\label{sparseconv}
\large
 f_{u,v}(\textbf{y}, \textbf{o}) = \frac{\sum_{i,j=-k}^{k} y_{u+i, v+j}o_{u+i, v+j}w_{i,j} }{\epsilon + \sum_{i,j=-k}^{k}o_{u+i, v+j}} + b
\end{equation}
where $o$ is a mask indicating the pixels that contain a valid depth value, $y$ is the input feature to the convolution, $w$ is the learned kernel, $k$ is the kernel size, $b$ is the bias, and $\epsilon > 0$ prevents division by 0. We choose a large kernel size of k=11 to have a big field of view before the encoder. The sparse convolution layers are followed by a ResNet-34 \cite{DBLP:journals/corr/HeZRS15} network, with group normalization layers replacing the usual batch normalization layers, because it has been shown in \cite{DBLP:journals/corr/abs-1803-08494} that they lead to a better generalization with smaller batch sizes, necessary for high resolution video training. Skip connections between encoder and decoder allow to transfer features from the previous frame to the new predicted frame in an implicit manner as shown in \cite{DBLP:journals/corr/abs-1708-06500}. Contrary to our baseline, we use convLSTMs \cite{10.5555/2969239.2969329} instead of LSTM to leverage more granular spatio-temporal features. The latent variables $z_{t}$ become tensors of size $3 \times 10 \times 512$, instead of $1 \times 1 \times 512$. Each $512$-vector in $z_{t}$ encodes a complex stochastic probability for a patch in the original $192 \times 640 $ image.\par

The loss function (Eq. \eqref{loss_depth}) for the $\mathbf{Depth}$ Network is a combination of a masked $L_{2}$ loss and a KL-divergence loss to optimize the variational lower bound as in \cite{DBLP:journals/corr/abs-1802-07687}.
\begin{equation}
\label{loss_depth}
\begin{split}
L_{\mathbf{Depth}} &=  \sum_{t} ( \sum_{pixels}|| \mathbbm{1}_{lidar(t)>0}.(dense(t) -lidar(t))  ||_{2}^{2} \\&+ \lambda_{1}\, L_{KL}(q_{\phi}(z_{t}|x_{1:t}), q_{\psi}(z_{t}|x_{1:t-1})) )
\end{split}
\end{equation}
Rather than utilizing the masked $L_{2}$ loss in the $\mathbf{Mask}$ network, instead a standard $L_{2}$ loss between the predicted mask and the groundtruth mask (which consists of all pixels that have a non-zero depth value) is employed. Although in recent works like SPCSF \cite{weng2020unsupervised}, the mask prediction is treated as a pixel classification problem, we have found that the $L_{2}$ loss is able to capture better details than the binary cross entropy loss. The latter leads to extremely sparse depth maps and a deterioration of the subsequent predicted frames.\par

To enable training on sequences consisting of 30 high resolution frames, we follow \cite{Patil_2020} by training with Truncated Back Propagation Through Time (TBPTT) \cite{tbptt}.  By updating the weights after each frame, less computations are required in comparison to Back Propagation Through Time and we were able to experiment with high capacity encoders. We trained with a batch size of 8 to prevent overfitting. Our training strategy consists in training the $\mathbf{Depth}$ Network and the $\mathbf{Mask}$ network each in an isolated way to predict the next frame. Then we trained the two networks together in an autoregressive manner by teacher forcing to predict 15 frames after being warmed up by 15 previous frames.
\begin{table}[t]
    \centering
    \caption{Quantitative comparisons}
    \label{table:nextframe}
    \begin{tabular}{ p{2.5cm} | p{1cm} p{1cm} p{1cm} p{1cm}  } 
    \hhline{=|===}
    & \multicolumn{3}{c}{(a) Next frame prediction} \\
     Framework      & MAE & RMSE &  CD  \\ 
    \hline
    (a) SVG-LP     & 5.36	 & 11.01 & 0.246        \\
    (b) Ours SLPC & \textbf{0.74} & \textbf{2.371} & \textbf{0.071}     \\
    \hhline{=|===}
     & \multicolumn{3}{c}{(b) Prediction of 15 frames} \\
    Framework      & MAE & RMSE &  CD  \\
    \hline
    (a) SPCSF  & 	16.01 & 20.76 &  1.135\\
    (b) SVG-LP  & 11.87	 & 17.30 & 0.963\\
    (c) Ours SLPC &\textbf{2.68} & \textbf{5.38} & \textbf{0.700} \\
    \hline
    \end{tabular}
    \vspace{-2mm}
\end{table}

\begin{figure}[t]
    \centering
    \includegraphics{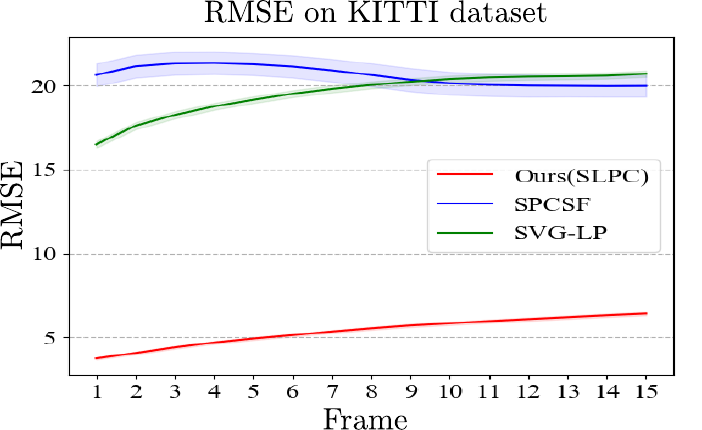}
    \caption{Quantitative analysis of our depth maps accuracy in comparison to the baselines. RMSE is plotted on 15 frames with 95\% confidence interval. All models are warmed up by 15 previous frames.}
    \label{fig:graphs}
    \vspace{-5mm}
\end{figure}  
\section{Datasets and Experiments}
\label{sec:experiments}
The proposed framework was evaluated on the KITTI dataset for depth completion \cite{Geiger2012CVPR}. It contains 85898 frames from 138 drives for training and around 7000 frames from 13 drives for validation. The LiDAR data was reprojected on a downscaled camera frame with a resolution of $192 \times 640$ using the given calibration data. We limited the videos to subsequences of 30 frames for training, and obtained 2554 training sequences in total. We trained on single NVIDIA TITAN RTX GPU.\par  
\begin{figure*}[h]
    \centering
    \includegraphics[width=\textwidth]{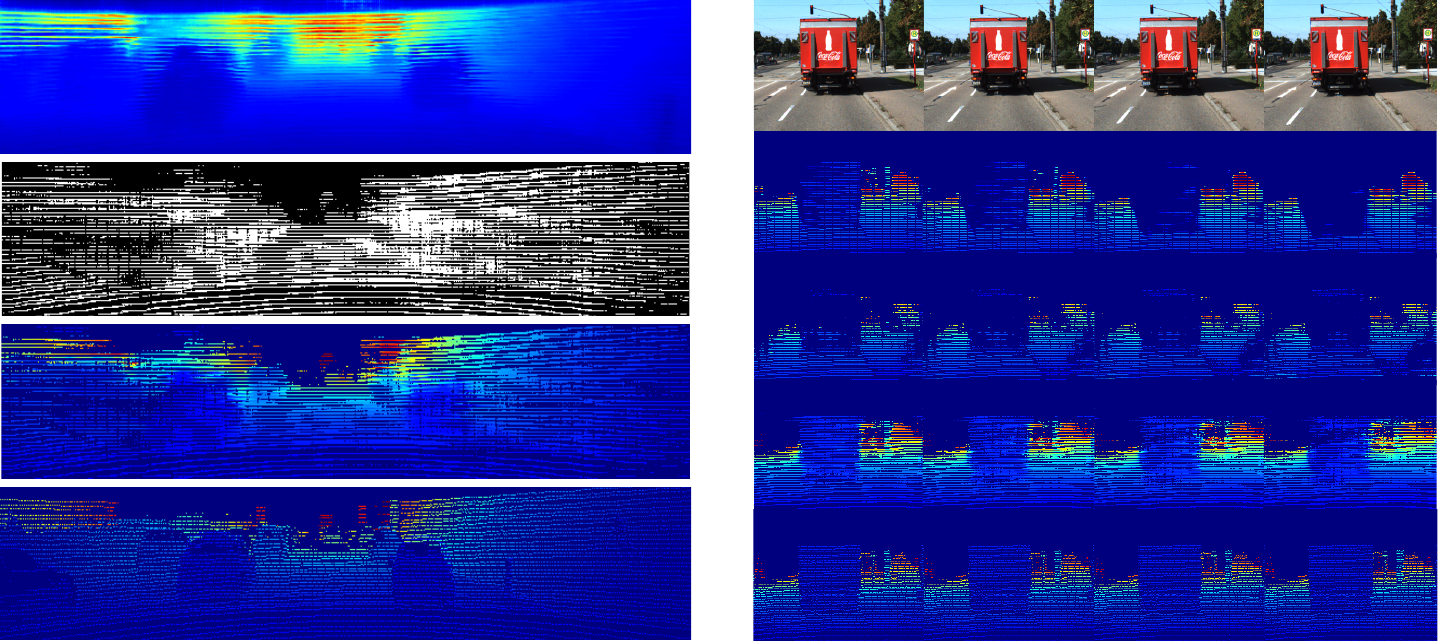}
    \caption{Qualitative Comparison.
    From top to bottom: Left: the predicted dense map, predicted mask, predicted LiDAR frame, groundtruth.
    Right: RGB images, SVG-LP, SPCSF, Ours, Groundtruth.  We illustrate 4 frames (17-20 in the sequence).}
    \label{fig:frames}
    \vspace{-4mm}
\end{figure*} 
Since there is no direct baseline, we compared with the SPCSF and the SVG-LP frameworks for Lidar Prediction. To achieve a fair comparison, we used the verified open-source implementation of SVG-LP while reproducing a faithful implementation of the architecture of SPCSF and using the original hyperparameters. We have performed two experiments: first, we have trained our $\mathbf{Depth}$ and $\mathbf{Mask}$ networks on the KITTI dataset for the task of next frame prediction and compared with SVG-LP (since SPCSF is trained to produce multiple frames at once). Then we trained the two networks together to predict 15 frames after being conditioned by 15 frames (which means 1.5 seconds of driving on the KITTI dataset). Comparisons were made with both SVG-LP and SPCSF. Since our model is stochastic, we draw 20 samples from the model for each testing sequence as in \cite{DBLP:journals/corr/abs-1802-07687} and report the mean RMSE for each predicted frame.\par

To evaluate the quantitative performance of our framework, the utilized metrics are the Root Mean Squared Error (RMSE) and Mean Absolute Error (MAE) in meters against the projected lidar frames, and the Chamfer Distance (CD) between the generated pointclouds and the original LiDAR pointclouds, projected in 3D using the camera calibration matrix in the KITTI dataset.
\section{Results and Discussion}
\label{sec:results}
The quantitative results in Table \ref{table:nextframe} and Fig. \ref{fig:graphs} indicate that our model outperforms the baselines by a large margin in all three metrics. Interestingly, the SPCSF model, although optimized to LiDAR prediction on range maps, performed worse than the SVG-LP baseline. We hypothesize, this is due to the nature of the depth maps: they are sparser than range maps and lie on the camera frame of reference as opposed to the reference LiDAR frame. We also offer the advantage of generating dense depth maps as seen in Fig. \ref{fig:frames}, which results in better accuracy. Moreover, SPCSF is deterministic as opposed to both SVG-LP and our model, which might lead to overfitting in some driving scenarios.\par
\vspace{-0.5mm}
The qualitative analysis in Fig. \ref{fig:frames} shows the robustness of our model against the sparse LiDAR. Both SVG-LP and SPCSF cannot simultaneously predict the depth and sparsity information accurately, and this resulted in depth maps with few points (or a lot of black holes in the depth estimation) especially in the latter frames. The mask network in our case prevents this degradation scenario from happening as it almost keeps the number of predicted points constant in each frame. In the case of SVG-LP, the resulting depth maps suffer from grid artifacts because the model was originally designed to deal with images not pointclouds. SPCSF and our Model do not suffer from this visual problem, because the loss functions of both models are more adapted to the 3D nature of LiDAR. However, the chamfer distance loss in SPCSF leads to pointclouds that are overpopulated in some locations and sparse in others, as was shown in \cite{Achlioptas2018LearningRA}. \par 
Our model however has some limitations. Most importantly is the blurry nature of our depth maps. This is attributed to the use of VAE as noted in \cite{DBLP:journals/corr/abs-1804-01523, DBLP:journals/corr/abs-1904-12165}. We will work in the near future to address this issue: visual quality can be enhanced by either adversarial methods \cite{DBLP:journals/corr/abs-1804-01523} or VAEs with hierarchical latent codes \cite{DBLP:journals/corr/abs-1904-12165}. Simple photometric losses from the depth completion literature such as in \cite{DBLP:journals/corr/abs-1806-01260,DBLP:journals/corr/abs-1807-00275} can be used for our self-supervised framework. Another limitation lies in our mask predictions, which do not have the same level of sparsity as the original LiDAR frames and lead to a deterioration in the prediction quality over time, as it can be seen in Fig. \ref{fig:graphs}. We plan in the near future to address these shortcomings.
\section{Conclusion}
In this work, we introduced the new task of depth map prediction and completion. To accomplish this, we presented a new framework and training algorithm. Our model SLPC provides robust depth prediction against sparse LiDAR. This was achieved by two VRNNs working in parallel, one to complete the depth information and the other to predict a sparsity mask that downsamples the predicted depth map. The architecture and losses are tailored to the nature of the data. A comparative analysis with state-of-the-art frameworks has illustrated the superior performance of our model for LiDAR prediction. Our method is orthogonal to previous frameworks in video prediction, meaning that any video prediction model can be combined with our method to produce state-of-the-art predicted depth maps.
\label{sec:conclusion}

\begingroup
\setstretch{0.80}
\setlength\bibitemsep{0pt}
\printbibliography

@inproceedings{DBLP:journals/corr/abs-1807-00275,
author = {Ma, Fangchang and Cavalheiro, Guilherme and Karaman, Sertac},
year = {2019},
pages = {3288-3295},
booktitle= {International Conference on Robotics and Automation {(ICRA)}},
title = {Self-Supervised Sparse-to-Dense: Self-Supervised Depth Completion from LiDAR and Monocular Camera},
}

@inproceedings{schreiber2019motion,
  author    = {Marcel Schreiber and
               others},
  title     = {Long-Term Occupancy Grid Prediction Using Recurrent Neural Networks},
  booktitle = {International Conference on Robotics and Automation {(ICRA)}},
  pages     = {9299--9305},
  year      = {2019},
}

@inproceedings{DBLP:journals/corr/FinnGL16,
  author    = {Chelsea Finn and
               Ian J. Goodfellow and
               Sergey Levine},
  title     = {Unsupervised Learning for Physical Interaction through Video Prediction},
  booktitle = {Conference on Neural Information Processing Systems {(NIPS)}},
  year      = {2016},

}

@inproceedings{DBLP:journals/corr/abs-1902-06919,
  author    = {Yafei Song and
               Yonghong Tian and
               Gang Wang and
               Mingyang Li},
  title     = {2D LiDAR Map Prediction via Estimating Motion Flow with {GRU}},
  booktitle = {International Conference on Robotics and Automation, {(ICRA)}},
  pages     = {6617--6623},
  year      = {2019},
}

@inproceedings{DBLP:journals/corr/abs-1802-07687,
  author    = {Emily Denton and
               Rob Fergus},
  title     = {Stochastic Video Generation with a Learned Prior},
  booktitle = {International Conference on Machine Learning,
               {(ICML)}},
  year      = {2018},
}

@article{DBLP:journals/corr/abs-1804-01523,
  author    = {Alex X. Lee and
               others},
  title     = {Stochastic Adversarial Video Prediction},
  journal   = {arXiv preprint},
  year      = {2018},
  url       = {http://arxiv.org/abs/1804.01523},
}

@inproceedings{DBLP:journals/corr/LotterKC16,
  author    = {William Lotter and
               Gabriel Kreiman and
               David D. Cox},
  title     = {Deep Predictive Coding Networks for Video Prediction and Unsupervised
               Learning},
  booktitle = {International Conference on Learning Representations, {(ICLR)}},
  year      = {2017},
}

@inproceedings{DBLP:journals/corr/AgrawalCM15,
  author    = {Pulkit Agrawal and
               Jo{\~{a}}o Carreira and
               Jitendra Malik},
  title     = {Learning to See by Moving},
  booktitle = {{IEEE} International Conference on Computer Vision {(ICCV)}},
  pages     = {37--45},
  year      = {2015},
  url       = {https://doi.org/10.1109/ICCV.2015.13},
}

@inproceedings{kingma2013auto,
  author    = {Diederik P. Kingma and
               Max Welling},
  title     = {Auto-Encoding Variational Bayes},
  booktitle = {International Conference on Learning Representations, {(ICLR)}},
  year      = {2014},
}

@inproceedings{goodfellow2014generative,
  title={Generative adversarial nets},
  author={Goodfellow, Ian and Pouget-Abadie, Jean and Mirza, Mehdi and Xu, Bing and Warde-Farley, David and Ozair, Sherjil and Courville, Aaron and Bengio, Yoshua},
  booktitle={Advances in neural information processing systems},
  pages={2672--2680},
  year={2014}
}

@inproceedings{DBLP:journals/corr/KalchbrennerOSD16,
  author    = {Nal Kalchbrenner and
               others},
 
  title     = {Video Pixel Networks},
  booktitle = {International Conference on Machine Learning,
               {(ICML)} },
  volume    = {70},
  pages     = {1771--1779},
  year      = {2017},

}

@inproceedings{DBLP:journals/corr/SrivastavaMS15,
  author    = {Nitish Srivastava and
               Elman Mansimov and
               Ruslan Salakhutdinov},
  title     = {Unsupervised Learning of Video Representations using LSTMs},
  booktitle = {International Conference on Machine Learning,
               {(ICML)}},
  series    = {{JMLR} Workshop and Conference Proceedings},
  volume    = {37},
  pages     = {843--852},
  year      = {2015},
}

@inproceedings{DBLP:journals/corr/OhGLLS15,
  author    = {Junhyuk Oh and
               others},
  title     = {Action-Conditional Video Prediction using Deep Networks in Atari Games},
  booktitle = {Conference
               on Neural Information Processing Systems {(NIPS)}},
  year      = {2015},
}

@inproceedings{rezende2014stochastic,
  author    = {Danilo Jimenez Rezende and
               others},
  title     = {Stochastic Backpropagation and Approximate Inference in Deep Generative
               Models},
  booktitle = {International Conference on Machine Learning,
               {(ICML)}},
  volume    = {32},
  pages     = {1278--1286},
  year      = {2014},
}

@inproceedings{DBLP:journals/corr/ChungKDGCB15,
  author    = {Junyoung Chung and
               others},
  title     = {A Recurrent Latent Variable Model for Sequential Data},
  booktitle = {Conference
               on Neural Information Processing Systems {(NIPS)}},
  year      = {2015},
}

@inproceedings{DBLP:journals/corr/abs-1904-12165,
  author    = {Llu{\'{\i}}s Castrej{\'{o}}n and
               others},
  title     = {Improved Conditional VRNNs for Video Prediction},
  booktitle = {International Conference on Computer Vision {(ICCV)}
               },
  year      = {2019},
}

@inproceedings{DBLP:journals/corr/abs-1708-06500,
   author    = {Jonas Uhrig and
               others},
  title     = {Sparsity Invariant CNNs},
  booktitle = {International Conference on {3D} Vision},
  pages     = {11--20},
  year      = {2017},
}

@article{DBLP:journals/corr/abs-1811-00684,
  author    = {Fitsum A. Reda and
               others},
  title     = {{SDCNet: Video Prediction Using Spatially-Displaced Convolution}},
  journal   = {European Conference on Computer Vision {(ECCV)}},
  year      = {2018},
}

@inproceedings{DBLP:journals/corr/abs-1904-12374,
  author    = {Masha Itkina and
               Katherine Rose Driggs{-}Campbell and
               Mykel J. Kochenderfer},
  title     = {Dynamic Environment Prediction in Urban Scenes using Recurrent Representation
               Learning},
  booktitle = {{IEEE} Intelligent Transportation Systems Conference {(ITSC)}},
  year      = {2019},
}

@inproceedings{10.5555/2969239.2969329,
author = {Shi, Xingjian and others},
title = {{Convolutional LSTM Network: A Machine Learning Approach for Precipitation Nowcasting}},
year = {2015},
booktitle = {Conference on Neural Information Processing Systems {(NIPS)}},
pages = {802–810},
}

@INPROCEEDINGS{Geiger2012CVPR,
  author = {Andreas Geiger and Philip Lenz and Raquel Urtasun},
  title = {Are we ready for Autonomous Driving? The KITTI Vision Benchmark Suite},
  booktitle = {{Conference on Computer Vision and Pattern Recognition (CVPR)}},
  year = {2012}
}

@inproceedings{DBLP:journals/corr/HeZRS15,
  author    = {Kaiming He and
               others},
  title     = {Deep Residual Learning for Image Recognition},
  booktitle = {{IEEE} Conference on Computer Vision and Pattern Recognition
               {(CVPR)}},
  pages     = {770--778},
  year      = {2016},
}

@article{DBLP:journals/corr/abs-1803-08494,
  author    = {Yuxin Wu and
               Kaiming He},
  title     = {Group Normalization},
  journal   = {International Journal of Computer Vision},
  volume    = {128},
  number    = {3},
  year      = {2020},
}

@misc{weng2020unsupervised,
    title={Unsupervised Sequence Forecasting of 100,000 Points for Unsupervised Trajectory Forecasting},
    author={Xinshuo Weng and others},
    year={2020},
    eprint={2003.08376},
    archivePrefix={arXiv},
}

@article{Patil_2020,
   title={Don’t Forget The Past: Recurrent Depth Estimation from Monocular Video},
   volume={5},
   journal={IEEE Robotics and Automation Letters},
   author={Patil, Vaishakh and others},
   year={2020},
   month={Oct},
   pages={6813–6820}
}

@article{tbptt,
author = {Williams, Ronald and Peng, Jing},
year = {1998},
month = {9},
title = {An Efficient Gradient-Based Algorithm for On-Line Training of Recurrent Network Trajectories},
journal = {Neural Computation},
}

@inproceedings{DBLP:journals/corr/FanSG16,
  author    = {Haoqiang Fan and
               Hao Su and
               Leonidas J. Guibas},
  title     = {A Point Set Generation Network for 3D Object Reconstruction from a
               Single Image},
  booktitle = {{IEEE} Conference on Computer Vision and Pattern Recognition {(CVPR)}},
  pages     = {2463--2471},
  year      = {2017},
}

@inproceedings{DBLP:journals/corr/abs-1806-01260,
  author    = {Cl{\'{e}}ment Godard and
               others},
  title     = {Digging Into Self-Supervised Monocular Depth Estimation},
  booktitle = {International Conference on Computer Vision {(ICCV)}},
  pages     = {3827--3837},
  year      = {2019},
}

@INPROCEEDINGS{rad1,
	author={Karim {Armanious} and others},
	booktitle={27th European Signal Processing Conference (EUSIPCO)}, 
	title={An Adversarial Super Resolution Remedy for Radar Design Trade-offs}, 
	year={2019},
	volume={},
	number={}
}

@INPROCEEDINGS{rad2,
	author={Sherif {Abdulatif} and others},
	booktitle={International Radar Conference (RADAR)}, 
	title={Towards Adversarial Denoising of Radar Micro-Doppler Signatures}, 
	year={2019},
	volume={},
	number={},
	pages={1-6}
	}

@INPROCEEDINGS{rad3,
	author={Sherif {Abdulatif} and others},
	booktitle={IEEE Radar Conference (RadarConf)}, 
	title={Person Identification and Body Mass Index: A Deep Learning-Based Study on Micro-Dopplers}, 
	year={2019},
	volume={},
	number={},
	pages={1-6}
}

@INPROCEEDINGS{rad4,
  author={Sherif {Abdulatif} and others},
  booktitle={2017 IEEE 6th Global Conference on Consumer Electronics (GCCE)}, 
  title={Stairs detection for enhancing wheelchair capabilities based on radar sensors}, 
  year={2017},
  volume={},
  number={},
  pages={1-5},}

@inproceedings{Achlioptas2018LearningRA,
  title={Learning Representations and Generative Models for 3D Point Clouds},
  author={Panos Achlioptas and O. Diamanti and Ioannis Mitliagkas and L. Guibas},
  booktitle={ICML},
  year={2018}
}
\endgroup

\end{document}